\documentclass[conference]{IEEEtran}
\IEEEoverridecommandlockouts
\usepackage{algorithm,algorithmic}
\usepackage{cite}
\usepackage{fancyhdr}
\usepackage{amsmath,amssymb,amsfonts}
\usepackage{algorithmic}
\usepackage{graphicx}
\usepackage{textcomp}
\usepackage{xcolor}
\usepackage{hyperref}
\usepackage{float}
\usepackage{fancyhdr}
\usepackage{lastpage}
\usepackage{eso-pic}
\usepackage{fancyhdr}
\usepackage{xurl}
\newlength\myindent
\newcommand\bindent{%
  \begingroup
  \setlength{\itemindent}{\myindent}
  \addtolength{\algorithmicindent}{\myindent}
}
\newcommand\eindent{\endgroup}
\def\BibTeX{{\rm B\kern-.05em{\sc i\kern-.025em b}\kern-.08em
    T\kern-.1667em\lower.7ex\hbox{E}\kern-.125emX}}
\begin{document}

\title{EWasteNet: A Two-Stream Data Efficient Image Transformer Approach for E-Waste Classification}

\author{Niful Islam, Md. Mehedi Hasan Jony, Emam Hasan, 
Sunny Sutradhar, Atikur Rahman, Md. Motaharul Islam\\
Department of Computer Science and Engineering, United International University\\ United City, Madani Avenue, Badda, Dhaka 1212, Bangladesh\\
\{nislam201057, mjony201336, ehasan201302, ssutradhar201084, arahman192087\}@bscse.uiu.ac.bd,\\ Corresponding author: motaharul@cse.uiu.ac.bd
}
\maketitle

\begin{abstract}
Improper disposal of e-waste poses global environmental and health risks, raising serious concerns. The accurate classification of e-waste images is critical for efficient management and recycling. In this paper, we have presented a comprehensive dataset comprised of eight different classes of images of electronic devices named the E-Waste Vision Dataset. We have also presented EWasteNet, a novel two-stream approach for precise e-waste image classification based on a data-efficient image transformer (DeiT). The first stream of EWasteNet passes through a sobel operator that detects the edges while the second stream is directed through an Atrous Spatial Pyramid Pooling and attention block where multi-scale contextual information is captured. We train both of the streams simultaneously and their features are merged at the decision level. The DeiT is used as the backbone of both streams. Extensive analysis of the e-waste dataset indicates the usefulness of our method, providing 96\% accuracy in e-waste classification. The proposed approach demonstrates significant usefulness in addressing the global concern of e-waste management. It facilitates efficient waste management and recycling by accurately classifying e-waste images, reducing health and safety hazards associated with improper disposal.

\end{abstract}


\section{Introduction}

Electronic waste, commonly referred to as e-waste, is a growing environmental issue caused by the disposal of electronic gadgets. Any appliance or electrical equipment that is powered by electricity or batteries, such as computers, cell phones, televisions, and kitchen appliances, falls under this category of waste \cite{ali2022transformer}. The swift progress of technology has increased the production of electronic devices, which has raised the generation of e-waste. Due to the toxic materials found inside these devices, improper management of e-waste can have a detrimental effect on both human health and the environment \cite{srivastav2023concepts, brindhadevi2023waste}. Recycling is one method of e-waste management. Since different electronic items are recycled differently, it is essential to classify the e-waste images correctly \cite{islam2023greener}. \par 
Convolutional Neural Networks (CNNs) have revolutionized the field of computer vision, allowing major developments in image recognition tasks \cite{jain2023optimized, majumdar2023gamma}. Although it has been very effective at classifying images, there are some drawbacks. Firstly, CNNs require an extensive amount of labeled training data to achieve optimal performance \cite{dua2023inception}. Such datasets can be time and money-consuming to collect and annotate. Additionally, traditional CNNs lack inherent scalability because growing the model's size frequently results in inefficient computation and higher memory needs \cite{bharadiya2023convolutional}. Finally, it has a limited ability to accommodate global context \cite{zhang2023vitaev2}. These constraints compelled researchers to investigate alternative architectures for image classification. \par 
Transformers initially intended for natural language processing, exhibited exceptional performance on various tasks. However, due to their innate sequential processing nature and lack of spatial inductive biases, applying transformers directly into images is complex. The Vision Transformer (ViT) was proposed as a solution to this problem by splitting an image into patches and treating them as tokens, allowing transformers to be applied to image classification \cite{dosovitskiy2020image}. Although ViT has produced impressive results, it has some drawbacks. It's reliance on a large number of patches causes increased memory consumption, limiting scalability. To address this shortcoming of both CNNs and ViT, data-efficient image transformers (DeiT) were developed \cite{touvron2021training}. \par 
In this paper, we present a dataset named E-Waste Vision Dataset that consists of 1053 images from eight classes (Mobile, TV, Laptop, keyboard, mouse, Microwave, Smartwatch, and Camera). Moreover, since the dataset has a limited number of samples, we employ a data-efficient image transformer (DeiT), a model capable of handling small datasets, for constricting a robust image classifier. The proposed architecture named EWasteNet has two streams (edge stream and pyramid stream). The edge stream consists of sobel operator that detects the edges and passes to the DeiT. The pyramid stream, on the other hand, incorporates Atrous Spatial Pyramid Pooling (ASPP) for capturing multi-scale information and Convolutional Block Attention Module (CBAM) for identifying and focusing the most important parts of the image. The features extracted from the streams are fused before the final classification. The model has been evaluated on a test dataset and has achieved an impressive accuracy of 96\%. Moreover, the EWasteNet consumes significantly fewer resources for training. To summarize, the paper has the following major contributions: 
\begin{itemize}
    \item We have proposed a dataset consisting of eight labels for e-waste classification.
    \item We have presented a novel architecture for e-waste classification that is made up of two streams of DeiT such as edge and pyramid.
    \item The proposed model achieves 96\% accuracy by consuming less than a million parameters with only twenty epochs of training. 
\end{itemize}

The subsequent sections are arranged in the following manner. Section \ref{sec:related-works} discusses the related works. Section \ref{sec:method} discusses the proposed system. Section \ref{sec:results} demonstrates the results and discussion. Finally, Section \ref{sec:conclusion} concludes the paper.
\section{Related Works}
\label{sec:related-works}
Authors have proposed various approaches for enhancing the efficiency and accuracy of electronic waste and materials classification. Cheng et al. \cite{cheng2022classification} proposed a siamese network-based model for classifying different types of electronic components. In the architecture, they incorporated a fine-tuned VGG16 with feature map concatenation as a feature extractor of siamese network. Moreover, they also presented a novel loss function named channel correlation loss for increasing the model's generalization ability. Nafiz et al. \cite{nafiz2023convowaste} presented a transfer learning-based approach where they fine-tuned a pre-trained InceptionResnetV2 and achieved 98\% accuracy. They integrated this model into an automatic waste segregation machine that can place the wastes into their corresponding bins. Atik \cite{atik2022classification} presented a custom CNN architecture made of six convolution layers and four pooling layers for electronic component classification. Kumsetty et al. \cite{kumsetty2022trashbox} presented a dataset named TrashBox made of seven classes. For classifying the images of that dataset, they used pre-trained CNN architecture for feature extraction and a quantum neural network for fine-tuning the feature vectors. For the CNN feature extractor, they experimented with five state-of-the-art image classifiers and found ResNet-101 performed the best. Kang et al. \cite{kang2020automatic} introduced multi-feature fusion, residual unit modification and activation function modification on a base ResNet-34 model to enhance the classification performance. In the multi-feature fusion stage, they used three parallel routes that extract features simultaneously and perform feature fusion at the end. The convolution kernel sizes vary in the three routes. In the updated residual block, instead of adding the output of the first convolutional layer to the next one, they used downsampling and passed it to the following layers. Finally, the improved ReLU activation function to overcome the dying ReLU problem. The new activation function has four input segments and depending on the input value, the calculation changes.  \par 
Liu et al. \cite{liu2022depth} enhanced the performance of ResNet-50 architecture by integrating a new attention module named Depth-wise Separable Convolution Attention Module (DSCAM). Inspired by the Convolution Block Attention Module (CBAM), the DSCAM module also includes channel attention and spatial attention. However, instead of point-wise convolution, the DSCAM block uses depth-wise separable convolution. They experimented with the model on four waste datasets and found the proposed model outperforms some well-known CNN architectures. This model, however, consumes huge resources while training. Another attention-based model with fewer parameters was proposed by Chen et al. \cite{chen2022garbage} where they introduced a parallel mixed attention mechanism (PMAM). PMAM computes channel attention and spatial attention parallelly which are later concatenated with the input features. While computing the channel attention, they merged the feature vectors obtained from global average pooling and global max pooling which is different from squeeze and excitation, the most popular channel attention mechanism \cite{sarker2023efficient}. The newly developed attention block was integrated with ShuffleNet V2. The model achieves an accuracy of 97\% on a custom dataset. \par 
For waste object detection, Cai et al. \cite{cai2022towards} presented a neural network architecture named YOLOG (YOLO for garbage detection). YOLOG improves YOLOv4 by integrating the DCSPResNet module that consists of residual connection and spatial pyramid pooling. Al Duhayyim et al. \cite{al2022deep} employed masked RCNN with ResNet-101 backbone for waste image detection. \par 
Farjana et al. \cite{iot4030011} proposed a system for collecting, classifying and ejecting e-wastes. They employed various sensors for locating e-wastes in an area which are then classified into two classes (metal and plastic). The metal wastes are recycled and plastic wastes are used to create bio-fuel and bio-chars. Shawpnil et al. \cite{shawpnil2023easye} ensembles different approaches for smart e-waste management. The system proposes dismantling electronic wastes by crushing them and differentiating the metallic part from the non-metallic parts with the help of the gravity separation technique from the trash. \par
Following the comparison, there remains a clear lack of adequate datasets for electronic waste image classification. Therefore, this paper presents a comprehensive dataset of images from eight classes for e-waste classification. We also present a two-stream DeiT-based approach for classifying the images. 

\section{Proposed System}
\label{sec:method}
\begin{figure}
    \centering
    \includegraphics[height=3cm,
    width=\linewidth]{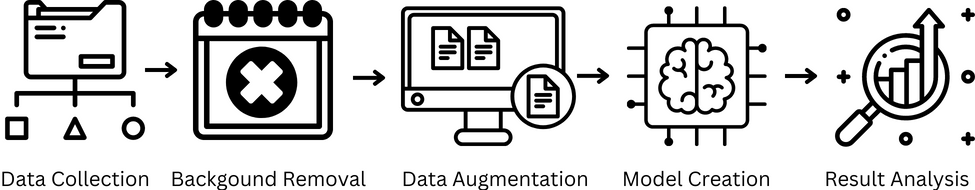}
    \caption{Workflow of the process}
    \label{fig:flowchart}
\end{figure}

This section describes the dataset along with the architecture of the proposed model. Figure \ref{fig:flowchart} presents a brief introduction of the whole process.
\begin{figure}[h]
    \centering
    \includegraphics[height=4.5cm,width=\linewidth]{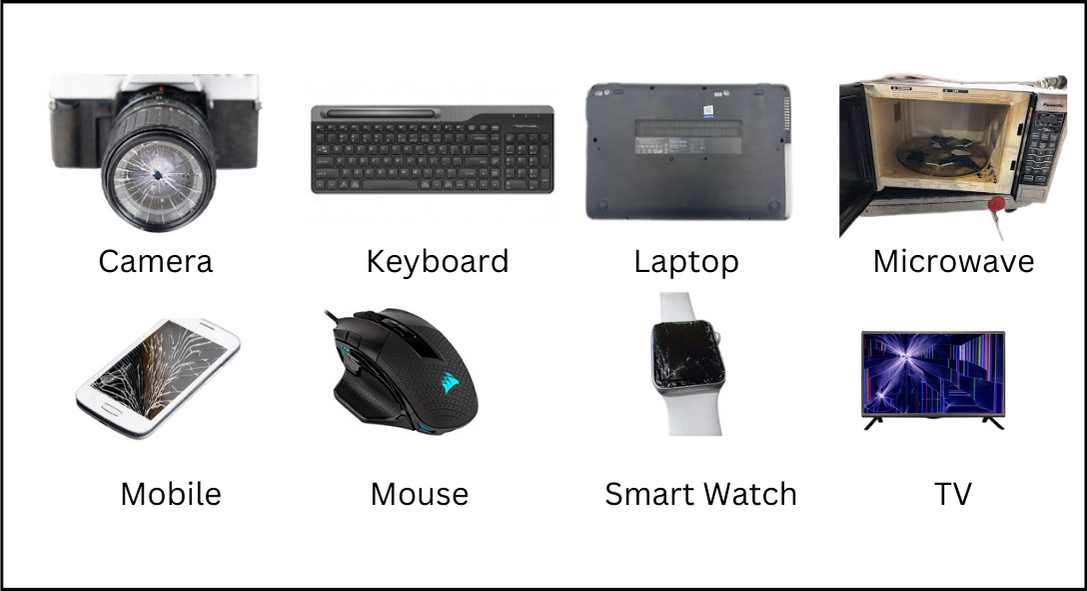}
    \caption{Sample of E-Waste Vision Dataset}
    \label{fig:dataset}
\end{figure}
\subsection{Data Collection and Preprocessing}
In this paper, we present the E-Waste Vision Dataset, which contains 1053 open-source images of various electronic devices such as mobile phones, televisions, laptop computers, keyboards, mouses, microwaves, smartwatches, and cameras. A sample of the dataset is presented in Figure \ref{fig:dataset}. We use a number of preprocessing steps to get the dataset ready for efficient training and evaluation. These include removing the background to isolate the objects of interest, rescaling the images to a standardized size of 348$\times$384 pixels, and augmenting the dataset with random rotation, height and weight shift, shear transformation, zooming, and flipping. In addition, we divided the preprocessed data into three sets: a training set, a validation set, and a test set, with respective ratios of 70:10:20. The data distribution is presented in Table \ref{tab:data}.
\begin{table}[!ht]
    \centering
    \caption{Distribution of E-Waste Vision Dataset}
    \begin{tabular}{|p{1.8cm}|p{1.2cm}|p{1.2cm}|p{1.2cm}|p{1.2cm}|}
    \hline
        \textbf{Class} & \textbf{Train} & \textbf{Validatioin} & \textbf{Test} & \textbf{Total} \\ \hline
        Camera & 60 & 2 & 16 & 78 \\ \hline
        Keyboards & 95 & 14 & 22 & 131 \\ \hline
        Laptop & 104 & 13 & 21 & 138 \\ \hline
        Microwave & 95 & 7 & 20 & 122 \\ \hline
        Mobile & 144 & 8 & 33 & 185 \\ \hline
        Mouses & 94 & 10 & 19 & 123 \\ \hline
        Smartwatch & 93 & 7 & 14 & 114 \\ \hline
        TV & 122 & 9 & 31 & 162 \\ \hline
        \textbf{Total} & \textbf{807} & \textbf{70} & \textbf{176} & \textbf{1053} \\ \hline
    \end{tabular}
    \label{tab:data}
\end{table}

\subsection{Proposed Architecture}
\begin{figure}[h]
    \centering
    \includegraphics[height=12cm, width=8cm]{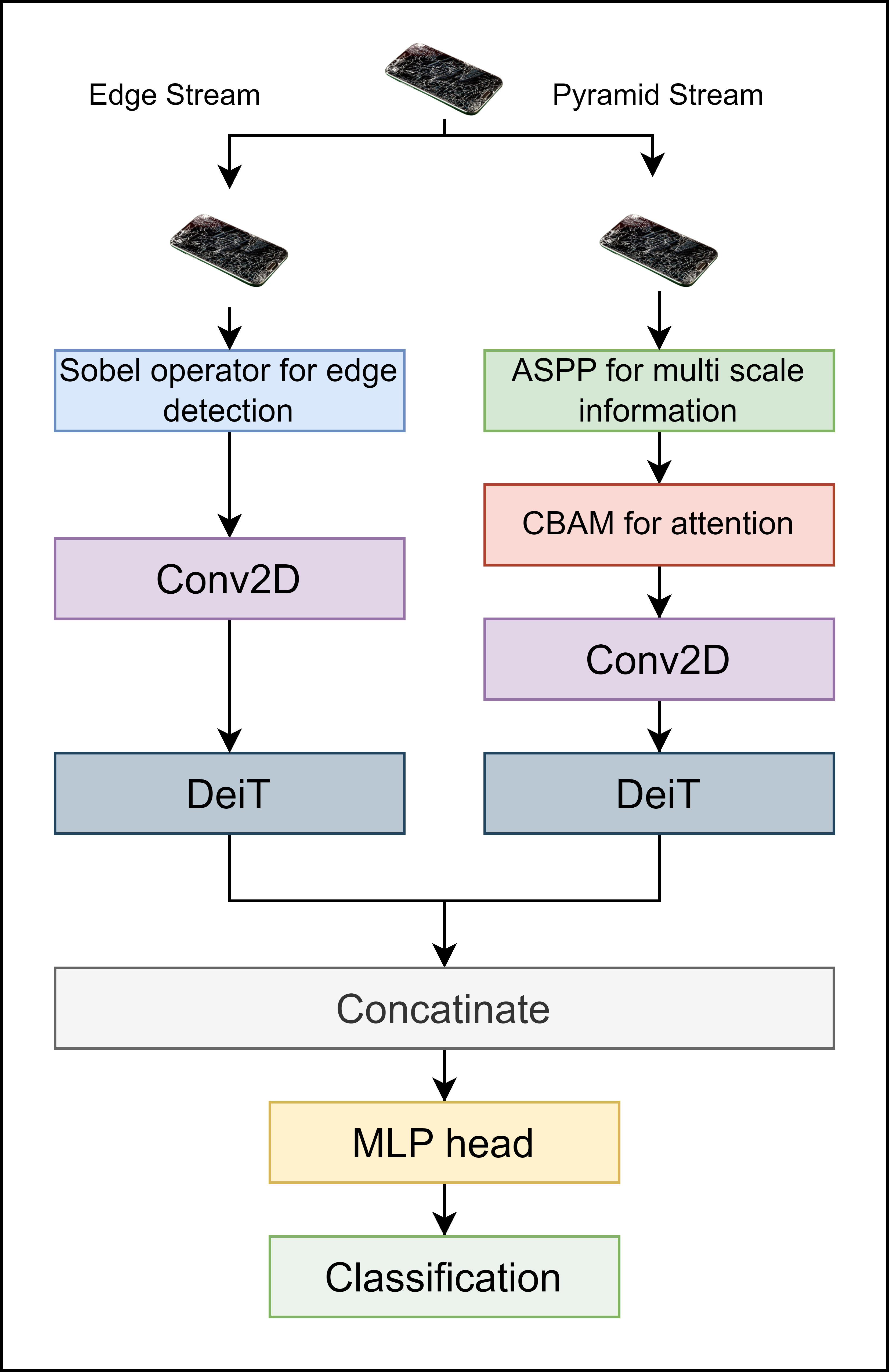}
    \caption{Proposed Architecture for E-Waste Classification}
    \label{fig:model}
\end{figure}

As shown in Figure \ref{fig:model}, the proposed model, EWasteNet, has a two streams DeiT architecture. The first stream is named as edge stream and the second stream as pyramid stream. 

\subsubsection{Edge Stream}
The importance of capturing refined edge information in image processing tasks necessitates the inclusion of an edge stream in a two-stream DeiT architecture. The sobel operator plays a crucial role in this stream by enabling the extraction of edge features from input images \cite{hao2022two}. The sobel operator's main task is to find edges by computing gradients at each pixel's location. Let the input image be defined with $I_m$ and output after sobel operation with $I$. Therefore, the operation carried out inside this module can be illustrated as shown in equation \ref{eq:sobel-operation}.

\begin{equation}
\label{eq:sobel-operation}
I=\begin{bmatrix}
-1 & 0 & 1 \\
-2 & 0 & 2 \\
-1 & 0 & 1
\end{bmatrix} * I_m
\end{equation}

A convolution layer is used in the edge stream after the sobel operator to match the input dimension of the pre-trained DeiT. The sobel outputs are convolved in this layer using a set of learned filters. The DeiT model, which uses self-attention mechanisms to capture global context and contextual relationships among edge features is finally introduced in the stream. 

\subsubsection{Prayed Stream}
In a two-stream DeiT architecture, a pyramid stream is required in order to efficiently capture multi-scale contextual information. By integrating features from various receptive field sizes, the pyramid stream improves the model's capacity to comprehend and categorize objects. This gives the model the flexibility to handle objects of different sizes and complexity levels, which is crucial for effective visual comprehension tasks. \par
\begin{figure}[h]
    \centering
    \includegraphics[height=7cm,width=\linewidth]{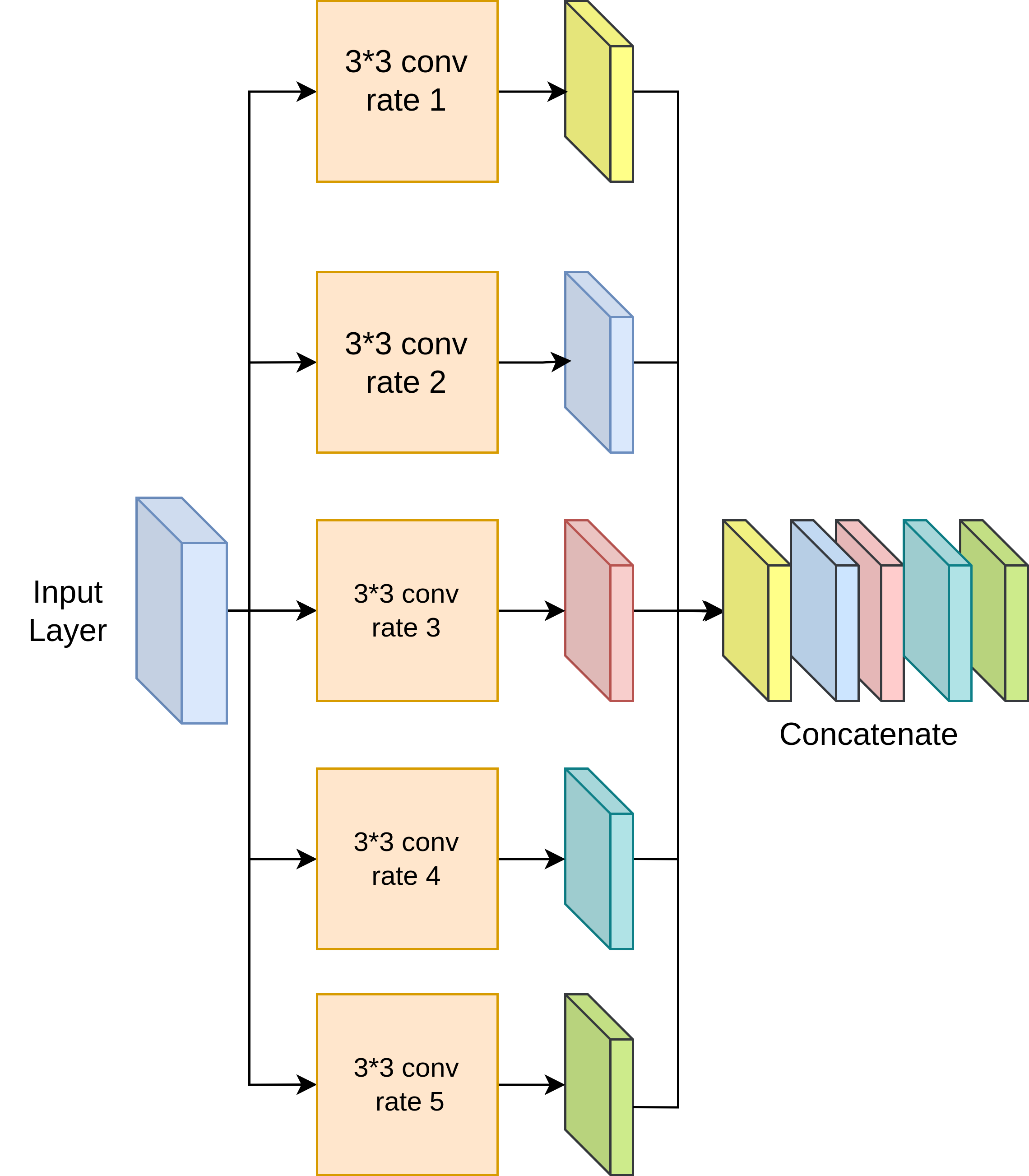}
    \caption{Atrous Spatial Pyramid Pooling for Proposed Model}
    \label{fig:aspp}
\end{figure}

 
The ASPP block, which accumulates multi-scale contextual data, is essential to the pyramid stream. By using parallel convolutional layers with various dilation rates, it accomplishes this. The dilation rates control the effective receptive field size of the convolutional filters, enabling them to acquire features at different scales. The ASPP block in our model is made up of five convolution layers, each with a unique dilation rate (1, 2, 3, 4, and 5). To ensure computational effectiveness, these layers are accompanied by decreasing numbers of filters (64, 32, 16, 8, and 4, respectively). The architecture of the ASPP block in our model is illustrated in Figure \ref{fig:aspp}. \par 

\begin{figure}[h]
    \centering
    \includegraphics[height=5cm,width=\linewidth]{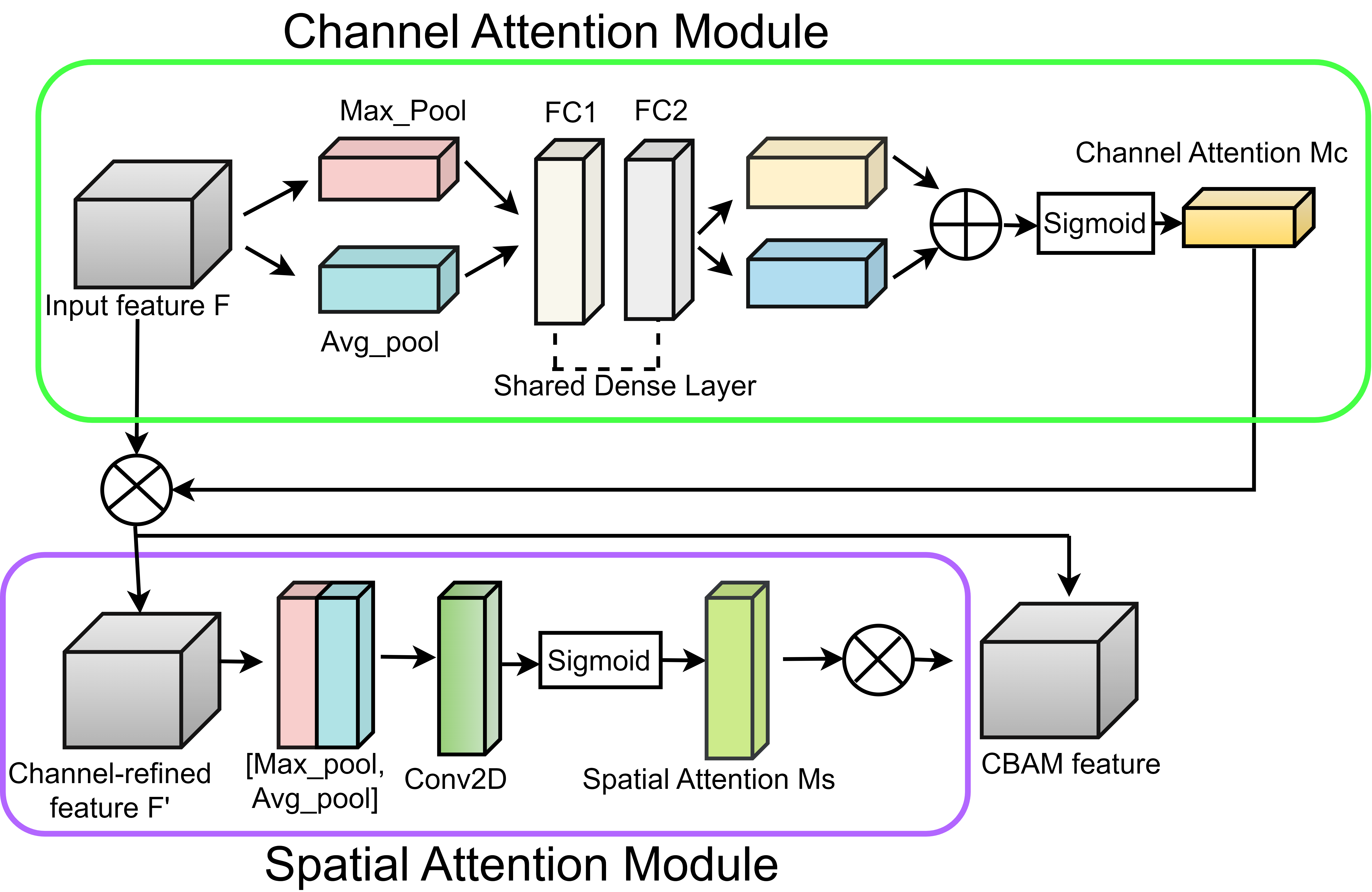}
    \caption{Convolutional Block Attention Module for Proposed Model}
    \label{fig:cbam}
\end{figure}

The importance of capturing both spatial and channel-wise attention leads to the demand for CBAM. The CBAM module improves the model's selective power by recalibrating feature maps adaptively \cite{cui2023maize}. As presented in Figure \ref{fig:cbam}, it accomplishes this by using a two-step procedure. First, by focusing on informative visual regions, the spatial attention mechanism captures fine-grained spatial dependencies. Second, the extraction of discriminative features is facilitated by the channel attention mechanism learning to emphasize pertinent channels. With the help of this dual attention mechanism, the model is better able to distinguish between various object classes since it efficiently captures both local and global contextual information.  Therefore the CBAM module is integrated after the ASPP block. \par 
Following the ASPP block and CBAM block, a convolution layer is again applied for matching the input dimension of DeiT. This layer also helps in fusing the multi-scale contextual information obtained from the ASPP block. Finally, the pre-trained DeiT model is integrated into the stream.

\subsubsection{Fusion of Two Streams}
In our research work, the two streams are fused to improve the overall classification performance by integrating their distinct features. These features are combined to create a unified representation after being extracted from the edge stream and the pyramid stream, respectively. This feature vector that has been concatenated is passed into a Multi-Layer Perceptron (MLP) for additional processing and classification.\par

A number of fully connected (FC) layers with distinct configurations make up the MLP architecture. A dropout layer with a 30\% dropout rate is used after an FC layer with 512 neurons. By randomly eliminating some of the neurons during training, this dropout layer aids in preventing overfitting. Following that, a fully connected layer with 256 neurons is inserted, followed by a 20\% dropout layer. To minimize the dimensionality of the feature representation, a bottleneck layer made up of 256 dense neurons is inserted last. The Rectified Linear Unit (ReLU) activation function is used to activate every FC layer in the MLP. ReLU contributes to the model's introduction of non-linearity, helping it to learn intricate patterns and representations. A softmax layer is used as the final classification layer. \par 

\begin{algorithm}[h]
\caption{EWasteNet Construction} 
\label{alg:algo}
\begin{algorithmic}
\STATE \textbf{Input :} H = Image height, W = Image Width
\STATE \textbf{Output :} M = EWastenet Model
\STATE \textbf{Method :}
\STATE Edge Stream:
\bindent
\STATE Input1 = InputLayer(heght=H, width=W)
\STATE edge\_img = sobel\_operator(Input1)
\STATE conv1 = Conv2D(edge\_img)
\STATE stream1= DeiT(conv1)
\eindent
\STATE Pyramid Stream:
\bindent
\STATE Input2 = InputLayer(heght=H, width=W)
\STATE aspp\_output = ASPP(Input2)
\STATE cbam\_output = CBAM(aspp\_output)
\STATE conv2 = Conv2D(cbam\_output)
\STATE stream2 = DeiT(conv2)
\eindent
\STATE Combine:
\bindent
\STATE merged = Concatenate(stream1, stream2)
\STATE mlp= MLP(merged)
\eindent
\RETURN  Model(input =[Input1, Input2] output=mlp)

\end{algorithmic}
\end{algorithm}
The construction process of the proposed model is further elaborated in Algorithm \ref{alg:algo}. The model construction function takes the image height and weight as input and returns the full model. The function first creates the endge stream followed by the pyramid stream. Later they are merged in the combine section. Finally the model is returned with two input layers from the two steams and the output layer being the softmax layer of the MLP head. \par 
One remarkable feature of the fusion strategy is its computational efficiency. The model strikes a balance between complexity and effectiveness with less than 1 million trainable parameters and only 20 epochs of training. The fusion method successfully merges the complementary data from the edge and pyramid streams, improving classification performance while retaining a manageable amount of parameters by utilizing the concatenated features and the MLP architecture. \par

\section{Results and Discussion}
\label{sec:results}
This section begins with a brief description of the evaluation metrics used to assess the model, followed by the results and subsequent discussion.

\subsection{Evaluation Matrices}

We used a number of evaluation criteria, including accuracy, precision, recall, F1-score, and Matthews Correlation Coefficient (MCC) to evaluate the effectiveness of our model for classifying e-waste \cite{islam2023distinguishing}. The model's capacity to correctly identify each class and strike a balance between precision and recall is illustrated as well by these metrics, which also offer insights into the total classification accuracy \cite{islam2023knntree}. For further understanding of the model's performance, we present the confusion matrix and the receiver operating characteristic (ROC) curve.

\begin{figure}[h]
    \centering
    \includegraphics[height=9cm,width=9cm]{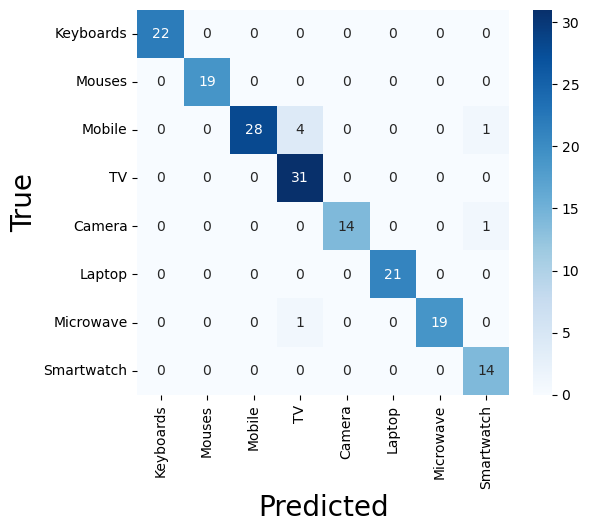}
    \caption{Confusion Matrix of the Proposed Model}
    \label{fig:cm}
\end{figure}

\subsection{Validation Results}
Our model attained an accuracy of 0.92 for the validation set. The proportion of accurately categorized instances among the expected positive examples as measured by the precision score was 0.93. In the same way, the recall score, which counts the percentage of occurrences that were correctly recognized among the actual positive examples was 0.93. The precision and recall-balancing F1-score was calculated to be 0.92. Finally, the MCC, which evaluates classification accuracy by accounting for true and false positives and negatives, produced a value of 0.91.

\begin{figure}[h]
    \centering
    \includegraphics[height=7cm, width=9cm]{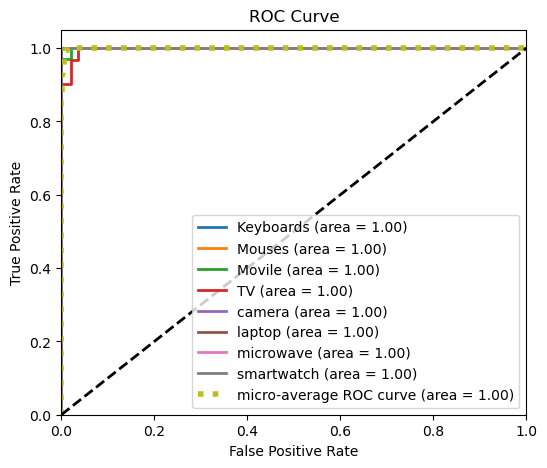}
    \caption{Receiver Operating Characteristic Curve of the Proposed Model}
    \label{fig:roc}
\end{figure}
\subsection{Test Results}
Our model's accuracy for the test set was even higher, at 0.96, demonstrating its sturdiness and generalizability. The precision score was calculated to be 0.96, indicating a high percentage of positively classified events that were correctly identified. The model's capacity to precisely identify real positive instances was demonstrated by the recall score of 0.9670. The F1-score was 0.96. Additionally, the test set's MCC score was 0.95, demonstrating a high degree of correlation between the predicted and actual labels. \par 
The confusion matrix provides specific information about how well the model classified each class. The confusion matrix of the model is presented in Figure \ref{fig:cm}. Based on the analysis of the confusion matrix, we observed that four occurrences of the mobile class were incorrectly classified as TV, one instance of the microwave class was incorrectly classed as TV, one instance of the mobile class was incorrectly classified as a smartwatch, and one instance of the camera class was incorrectly classified as a smartwatch. Aside from these errors in categorization, the model showed overall correct predictions for the remaining occurrences across all classes. \par 
The ROC curve is a graphical illustration of the trade-off between true positive rate (TPR) and false positive rate (FPR) at various classification thresholds. The ROC curve of our model is shown in Figure \ref{fig:roc}. For the test dataset, all classes had ROC scores of 1, suggesting exceptional performance in differentiating between positive and negative instances. Additionally, the micro-average ROC score, which takes into account the overall performance across all classes, reached the highest possible value of 1. Overall, the evaluation metrics and outcomes show that our algorithm is accurate and effective at classifying e-waste images. The reliability of our model and its potential for use in practical e-waste management applications is demonstrated by its high accuracy, precision, recall, F1 score, and MCC values.

\section{Conclusion}
\label{sec:conclusion}
This paper presents a new dataset for e-wast images along with a detailed study on the classification of e-waste images using a novel two-stream approach based on a Data-efficient Image Transformer (DeiT) architecture. The primary objective of the research was to accurately categorize electronic devices for effective waste management and recycling in order to address the growing concern over inappropriate e-waste management on a global scale. For facilitating this research, we created the E-Waste Vision Dataset, which contains 1053 open-source photos from eight distinct classes, including Mobile, TV, Laptop, keyboard, mouse, Microwave, Smartwatch, and Camera. A thorough analysis of the presented dataset showed the efficacy of our approach. On the validation and test sets, the model obtained excellent scores of accuracy, precision, recall, F1 score, and MCC, demonstrating its resilience and generalizability. Additionally, the reliability of the model was supported by the confusion matrix and ROC curve analysis. Overall, by offering an innovative approach to classification and a sizable dataset, our research advances the field of e-waste management. The outstanding accuracy and efficiency of the suggested model show promise for real-world applications, assisting in the ethical disposal and recycling of electronic trash.

\section*{Supplementary Materials}
\label{sec:suppli-materials}
The dataset along with the experiment can be found in this GitHub repository: \url{https://github.com/NifulIslam/EWasteNet-A-Two-Stream-DeiT-Approach-for-E-Waste-Classification}.

\bibliographystyle{IEEEtran}
\bibliography{FILES/bibfile.bib}

\end{document}